\documentclass[letterpaper]{article} % DO NOT CHANGE THIS
\usepackage{aaai2026}  % DO NOT CHANGE THIS
\usepackage{times}  % DO NOT CHANGE THIS
\usepackage{helvet}  % DO NOT CHANGE THIS
\usepackage{courier}  % DO NOT CHANGE THIS
\usepackage[hyphens]{url}  % DO NOT CHANGE THIS
\usepackage{graphicx} % DO NOT CHANGE THIS
\usepackage{natbib}  % DO NOT CHANGE THIS AND DO NOT ADD ANY OPTIONS TO IT
\usepackage{caption} % DO NOT CHANGE THIS AND DO NOT ADD ANY OPTIONS TO IT
\usepackage{algorithm}
\usepackage{algorithmic}
\usepackage{amsmath}
\usepackage{amsfonts}
\usepackage{booktabs}
\usepackage{multirow}
\usepackage{subcaption}

\usepackage{newfloat}
\usepackage{listings}
\DeclareCaptionStyle{ruled}{labelfont=normalfont,labelsep=colon,strut=off} % DO NOT CHANGE THIS
\floatstyle{ruled}
\newfloat{listing}{tb}{lst}{}
\floatname{listing}{Listing}
\title{\textsc{ID-\textsc{PaS}+}: Identity-Aware Predict-and-Search\\ for Solving General Mixed-Integer Linear Programs}
\author{
    Junyang Cai\textsuperscript{\rm 1},
    El Mehdi Er Raqabi\textsuperscript{\rm 2},
    Pascal Van Hentenryck\textsuperscript{\rm 2},
    Bistra Dilkina\textsuperscript{\rm 1}
}

\affiliations{
    \textsuperscript{\rm 1}University of Southern California, Los Angeles, CA, USA\\
    \textsuperscript{\rm 2}Georgia Institute of Technology, Atlanta, GA, USA\\
    \{caijunya, dilkina\}@usc.edu, \{eraqabi6, pvh\}@gatech.edu
}

\usepackage{bibentry}
\begin{document}
\maketitle              % typeset the header of the contribution
\begin{abstract}
Mixed-Integer Linear Programs (MIPs) are powerful and flexible tools for modeling a wide range of real-world combinatorial optimization problems. Predict-and-Search methods operate by using a predictive model to estimate promising variable assignments and then guiding a search procedure toward high-quality solutions. Recent research has demonstrated that incorporating machine learning (ML) into the Predict-and-Search framework significantly enhances its performance. Still, it is restricted to binary-only problems and overlooks the presence of fixed variable structures that commonly arise in real-world settings. This work extends the current Predict-and-Search (\textsc{PaS}) framework to parametric general parametric MIPs and introduces \textsc{ID-PaS+}, an identity-aware learning framework that enables the ML model to handle heterogeneous variable types more effectively. Experiments on several real-world large-scale problems demonstrate that \textsc{ID-PaS+} consistently achieves superior performance compared to the state-of-the-art solver Gurobi and \textsc{PaS}.

% \keywords{Discrete Optimization \and Deep Learning \and Mixed Integer Linear Programming\and Graph Neural Networks}
\end{abstract}
\section{Introduction}
This paper considers Parametric Mixed Integer Linear Programs (MIPs) of the form $P_I = (A, b, c)$ where we optimize for
\[
\min\{c^\top x \mid Ax \le b,\; x \in \mathbb{R}^n,\; x_j \in \mathbb{Z}\ \forall j \in I\},
\]
where $P$ is the instance data, $A \in \mathbb{R}^{m \times n}$, $b \in \mathbb{R}^m$, $c \in \mathbb{R}^n$, and $I \subseteq \{1,\ldots,n\}$ denotes the index set of general integer variables. The objective is to minimize $c^\top x$ subject to the linear constraints. Many such real-world MIPs contain millions of variables and constraints, making exact optimization challenging in environments where they must be solved quickly. However, in many industrial settings, these MIPs are repeatedly solved for instances drawn from distributions that are learned from historical data and/or forecasts. Machine learning (ML) offers an opportunity to reduce the search space of these MIPs by learning patterns from historical optimal or high-quality solutions. 

Recent work in this area has integrated ML with solvers, for example, through large neighborhood search~\cite{song2020general,huang2023searching,cai2024balans}, optimization proxies~\cite{chen2023end,tanneau2024dual}, or predict-then-optimize pipelines~\cite{tang2024pyepo,wu2024towards,mohan2025fairorml}. A complementary line of research develops ML-based methods that generate high-quality feasible solutions. The Predict-and-Search (\textsc{PaS}) framework~\cite{han2023gnn}, inspired by trust-region methods, explores a local neighborhood around a predicted partial assignment rather than fixing variable values outright, often yielding better solutions and reducing the risk of infeasible solutions. Recent improvements, such as contrastive learning~\cite{huang2024contrastive} and multitask learning~\cite{cai2024multi}, further enhance \textsc{PaS} performance. Yet, these methods remain largely limited to binary problems, whereas real-world MIPs frequently involve general integer variables~\cite{greening2023lead,akhlaghi2025propel,kim2025practice}. In addition, prior evaluations focus mainly on synthetic rather than large-scale industrial benchmarks.

Careful analysis of real-world Parametric MIPs---such as those governing power systems~\cite{bani2025decomposition}, supply chains~\cite{erraqabi5}, planning~\cite{cai2025neuro}, scheduling~\cite{ye2025cornell}, and rail networks~\cite{kim2025practice}---reveals two critical, shared properties. First, a substantial fraction of their variables take a value of exactly \textit{zero} in optimal or high-quality solutions~\cite{erraqabi1}. Second, because they model slowly changing physical infrastructures, the \textit{identity} of the variables remains consistent across instances (e.g., a specific variable always represents the same factory or train route, even as daily demand fluctuates)~\cite{akhlaghi2025propel}. 

The first observation is the key to extending \textsc{PaS} to general integer optimization. Predicting the exact value of a general integer variable is computationally inefficient and highly susceptible to error. However, by shifting the focus specifically to predicting which variables are \textit{exactly zero}, we can drastically prune the search space with high confidence. A non-zero value in real-world settings typically represents an active, costly decision (e.g., opening a facility or allocating a discrete batch of resources); consequently, the vast majority of available options remain unselected. Focusing on zero-values not only improves the model's accuracy by targeting the most common state, but it also provides the exact mechanism needed to generalize the \textsc{PaS} framework from strictly binary settings to general integer problems.

\begin{figure*} [ht]
    \centering
    \includegraphics[width=\linewidth]{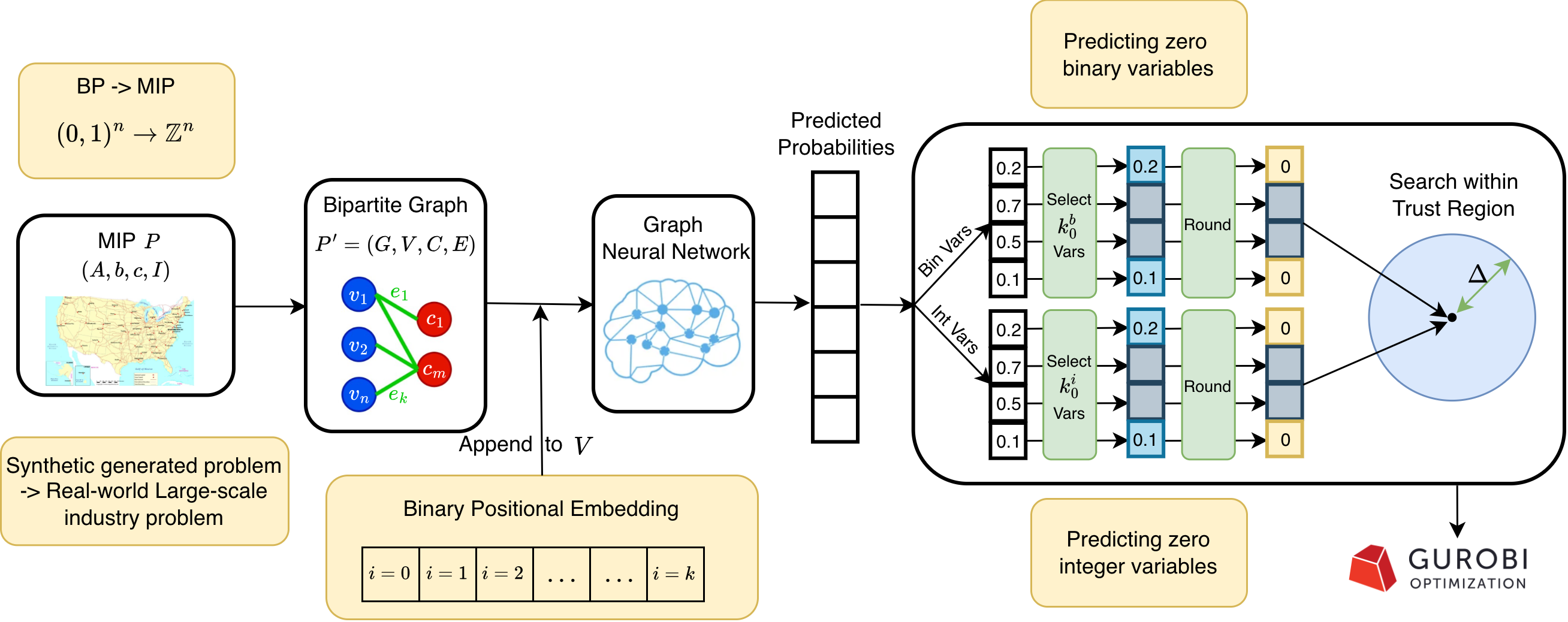}
    \caption{Overview of \textsc{ID-PaS+} for general MIPs: A MIP instance is represented as a bipartite graph with identity-aware positional embeddings. A GNN predicts the probability that each variable is zero, then the selection process splits into two distinct, parallel tracks for binary and integer variables. High-confidence variables from both the binary ($k_0^b$) and integer ($k_0^i$) sets are selected, rounded, and fixed to zero. A subsequent search within a trust region is then performed before passing the reduced problem to a solver. The components highlighted in yellow denote the key extensions over \textsc{PaS}: binary positional embeddings (identity awareness), specific predictions for zero-valued binary and integer variables, and the generalization from synthetic binary problems to large-scale, real-world industry MIPs.}
    \label{fig:id-pas}
\end{figure*}

Equally critical is the second observation regarding variable identity, which is essential for cross-instance generalization. Standard Graph Neural Networks (GNNs) represent MIPs as bipartite graphs and operate under permutation invariance. Consequently, they fail to recognize that ``variable 1'' in today's instance structurally corresponds to ``variable 1'' in tomorrow's instance. In real-world parametric MIPs, retaining this identity mapping is highly informative, as certain physical components consistently exhibit similar historical behaviors. By injecting identity-aware positional embeddings into the graph formulation, the model can learn and leverage these variable-specific historical priors, vastly improving its predictive accuracy on unseen instances drawn from the same underlying infrastructure.

Building on these two insights, this paper proposes \textsc{ID-PaS+}, a simple yet effective identity-aware framework that successfully bridges the gap between binary-only problems and general integer MIPs. \textsc{ID-PaS+} generalizes the \textsc{PaS} approach to general MIPs by smartly predicting zero-valued variables, avoiding the combinatorial burden of predicting all integer values. Compared to standard permutation-invariant graph networks, \textsc{ID-PaS+} incorporates identity-aware features to improve cross-instance generalization. Evaluated on three large-scale, real-world industry problems, \textsc{ID-PaS+} demonstrates consistent and significant performance improvements over both standard Gurobi and the original \textsc{PaS} framework.

\section{Background}
\subsection{Parametric MIP}
Many early works focus on parametric linear programs~\cite{gal2009linear}, where solution behavior under parameter changes is well understood. Parametric MIP, by contrast, concerns families of mixed-integer problems whose objectives or constraints depend on varying parameters, enabling analysis of how optimal solutions evolve across related instances. Parametric structure also plays a central role in areas such as online learning and control~\cite{russo2023learning,bertsimas2022online}. Recent progress within mixed-integer optimization has focused mainly on quadratic or nonlinear formulations~\cite{ranjan2025verification,pangia2025branch}, while general parametric linear problems remain far less explored.
 
\subsection{Predict-and-Search}
Predict-and-Search~\cite{han2023gnn} is a primal heuristic that leverages the prediction of the optimal solutions to guide the search process. Given a MIP instance $P_I$, let \(p_\pi(x_i \mid P_I)\) denote the predicted probability for each binary variable $x_i$. \textsc{PaS} identifies near-optimal solutions by exploring a neighborhood informed by these predictions. Specifically, it selects \(k_0\) binary variables \(X_0\) with the smallest \(p_\pi(x_i \mid P_I)\) and \(k_1\) binary variables \(X_1\) with the largest \(p_\pi(x_i \mid P_I)\), ensuring \(X_0\) and \(X_1\) are disjoint (\(k_0 + k_1 \leq q\)). Variables in \(X_0\) are fixed to 0, and those in \(X_1\) are fixed to 1 in a sub-MIP. However, \textsc{PaS} allows up to \(\Delta \geq 0\) of these fixed variables to be flipped during solving. \\
Formally, let 
\[
B(X_0, X_1, \Delta) = \{x : \sum_{x_i \in X_0} x_i + \sum_{x_i \in X_1} (1 - x_i) \leq \Delta \}
\]
be the neighborhood defined by \(X_0\), \(X_1\), and \(\Delta\), and let \(D\) represent the feasible region of the original MIP. \textsc{PaS} then solves the following optimization problem:
\[
\min c^T x \quad \text{s.t.} \quad x \in D \cap B(X_0, X_1, \Delta).
\]
Restricting the solution space to \(D \cap B(X_0, X_1, \Delta)\) simplifies the problem, enabling the solver to find high-quality feasible solutions to the original MIP more efficiently.

\section{The Identity-aware \\ General \textsc{PaS} Framework}

This paper extends the Predict-and-Search (\textsc{PaS}) framework, originally designed for binary MIPs, to general MIPs containing both binary and general integer variables. Inspired by~\cite{akhlaghi2025propel}, the core insight is to learn whether each discrete variable is zero, rather than attempting to predict its exact integer value. This strategy preserves compatibility with the trust-region search of \textsc{PaS} while directly exploiting the sparsity structure prevalent in real-world applications, where a vast majority of active decisions default to zero. Furthermore, this framework aligns with the reality of industrial settings: instances often share the exact same underlying variables and physical constraints, differing only in their parameters. \textsc{ID-PaS+} leverages this by encoding variable identity, allowing the model to exploit consistent historical semantics across different instances. The entire pipeline is illustrated in Figure \ref{fig:id-pas} and in the following subsections, we introduce these modules in detail.

\subsection{Data Collection}
For a given MIP $P_I$, the learning task collects a set of optimal or near-optimal solutions as training samples. Up to $u_p$ distinct solutions with best objective values are extracted (in our experiments, $u_p = 50$). Because the primary objective of the model is to identify zero-valued variables, \textsc{ID-PaS+} binarizes each training solution by keeping the zero entries as $0$ and converting all non-zero entries (regardless of their exact integer value) to $1$. This binarized representation provides a consistent, simplified learning target across both binary and general integer variables, drastically reducing the combinatorial complexity of the learning task.

\subsection{Bipartite Graph Representation}
Each MIP $P_I = (A, b, c)$ is represented as a bipartite graph following~\cite{gasse2019exact,cai2024learning}. The graph $P' = (G, V, C, E)$ contains variable nodes $V$ and constraint nodes $C$, with an edge $(i,j) \in E$ existing whenever variable $i$ appears in constraint $j$ with a non-zero coefficient. This representation uses 15 variable features (e.g., variable type, objective coefficient, bounds), 4 constraint features (e.g., right-hand side and sense), and 1 edge feature (the constraint matrix coefficient).

\textbf{Overcoming Permutation Invariance with Identity Awareness:} While standard bipartite encodings are beneficial for synthetic benchmarks, they are inherently permutation-invariant. Real-world problems, however, are typically instances of a parametric MIP governing a fixed underlying infrastructure (e.g., a specific power grid or railway network). Standard graph networks cannot recognize that a specific node always corresponds to the same physical entity across different daily instances. To address this critical limitation, \textsc{ID-PaS} introduces \textit{identity-aware features}. By appending a binary positional vector that encodes the unique index of each variable to the original feature set, \textsc{ID-PaS+} breaks permutation invariance. This preserves the structural advantages of GNNs while enabling the model to recognize and learn from the historical behavior of specific variable identities across instances.

\subsection{Model Architecture and Training Objective}
\textsc{ID-PaS+} learns a policy $\pi$ parameterized by a Graph Attention Network (GAT)~\cite{brody2021attentive}. The network takes the bipartite graph $P'$ as input and outputs a probability score for each variable. First, embedding layers map the node and edge features into a shared embedding dimension of size $L$. The GAT then performs two rounds of message passing: in the first round, each constraint node attends to its neighboring variable nodes using $H$ attention heads; in the second round, variable nodes attend back to constraints with a separate set of attention parameters. The convolution and attention parameters are shared across the graph, scaling the model size as $\Theta(HL^2)$. A multi-layer perceptron followed by a sigmoid activation then produces a probability score in $[0,1]$ for each variable. In our experiments, we set the embedding dimension $L=64$ and the number of attention heads $H=8$.

\textsc{ID-PaS+} trains the network using an imitation-learning objective~\cite{han2023gnn}. Given the binarized target vector $y \in \{0,1\}^{|I|}$ and the predicted probabilities $\hat{y} \in [0,1]^{|I|}$, where $I$ is the index set of all discrete (binary and integer) variables, \textsc{ID-PaS+} optimizes a binary cross-entropy loss~\cite{mao2023cross} over a batch of MIP instances $\mathcal{B}$:
\begin{align} \nonumber
\mathcal{L}(\theta) = - \sum_{P_I \in \mathcal{B}} \sum_{u \in u_p(P_I)} \sum_{i \in I} \Big( & y^u_i \log \hat{y}_i(\theta, P_I) \\
& + (1 - y^u_i) \log (1 - \hat{y}_i(\theta, P_I)) \Big). \nonumber
\end{align}
This simple yet effective loss encourages the network to assign the lowest possible probabilities to variables that consistently take the value of zero in high-quality solutions.

\subsection{Applying the Learned Network}
During inference, a new MIP instance is converted into a bipartite graph and evaluated by the GAT to produce continuous probability scores. In contrast to the original \textsc{PaS} framework~\cite{han2023gnn}, which attempts to predict both zero binaries and one binaries, \textsc{ID-PaS+} strictly focuses on zero-predictions and splits the outputs into two parallel sets: one for binary variables and one for general integer variables. The framework processes the scores to identify the variables most likely to be zero. Specifically, it selects the $k_0^b$ binary variables and the $k_0^i$ general integer variables that have the lowest predicted probabilities of being non-zero. These selected variables are rounded and form a combined set $X_0$, where $|X_0| = k_0^b + k_0^i$. The variables in $X_0$ are initially fixed to $0$. 

To prevent the solver from being trapped by inaccurate predictions, similar to \textsc{PaS}, \textsc{ID-PaS+} also defines a trust region that allows the solver to flip up to $\Delta$ of these fixed variables during optimization. The search is thus restricted to the \textsc{PaS} neighborhood:
\[
B(X_0, \Delta) = \Big\{ x : \sum_{x_i \in X_0} \mathbf{1}[x_i \neq 0] \le \Delta \Big\}.
\]
The reduced MIP is then solved over the feasible region intersected with $B(X_0, \Delta)$. By treating binary and integer zero-predictions in parallel, \textsc{ID-PaS+} provides highly reliable guidance that drastically prunes the search space while remaining robust to occasional misclassifications.

\begin{figure} [t]
    \centering
    \includegraphics[width=\linewidth]{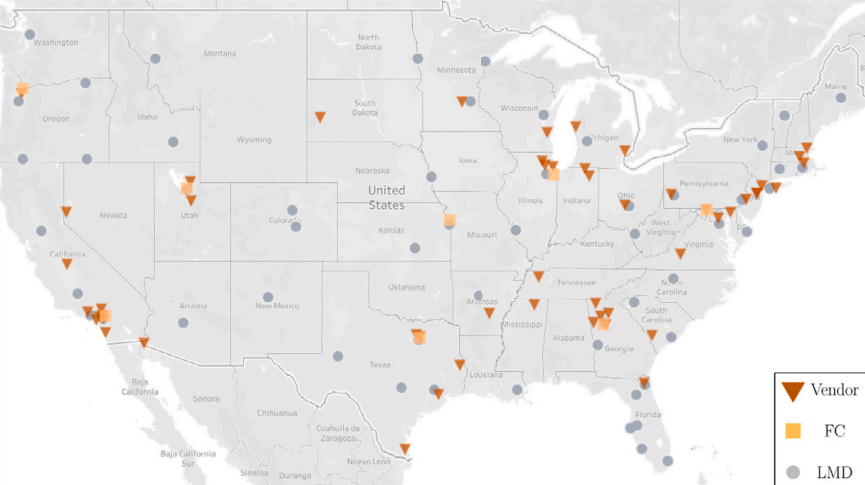}
    \caption{Middle-Mile Consolidation Network Design: example from \cite{greening2023lead}.}
    \label{fig:fullnetworkmmcnp}
\end{figure}

\section{Real-world Benchmarks}

\subsection{Middle-Mile Consolidation Network Design}
The first set of experiments considers the middle-mile consolidation network design problem (MMCNP)~\cite{greening2023lead,huang2024distributional}. The MMCNP focuses on planning transportation capacity from vendors to fulfillment centers (FCs) and last-mile delivery (LMDs) centers within required lead times (see Figure~\ref{fig:fullnetworkmmcnp} for an example). Shipments must be routed and consolidated into scheduled loads to ensure efficient delivery. The planner must decide which connections to operate, how much capacity to allocate, and how to group shipments so all demand is delivered on time. The large problem setting includes 50 facilities, 2{,}000 paths, 400 arcs, and 300 commodities, and the very large (designated as \emph{-Hard}) problem setting includes 200 facilities, 24{,}000 paths, 1{,}000 arcs, and 6{,}000 commodities. In the large setting, instances close the optimality gap in an average of 18 hours. In the very large setting, instances cannot converge and have an average gap of 5.12\% after 24 hours, underscoring the difficulty of these models.

We assumes that what varies across problem instances is the demand for each commodity. To generate instances, the experiments vary each commodity demand around its reference value using a normal distribution with bounds to keep demands realistic relative to arc capacities. Crucially, while daily demand fluctuates, the underlying physical infrastructure (the specific facilities and available transportation arcs) remains entirely fixed across instances. This static topology means that mathematical variables consistently represent the exact same physical connections, retaining a persistent identity characterized by historical routing behaviors. In this model, the general integer variables capture the number of trucks assigned to each arc. Since real consolidation networks operate sparsely, with only a highly optimized fraction of all possible connections used, the vast majority of these integer variables are exactly zero in high-quality solutions. Focusing on the underlying sparsity of these zero-valued integer variables provides a pathway to bypass the combinatorial complexity of determining exact truck counts.

\begin{figure} [t]
    \centering
    \includegraphics[width=\linewidth]{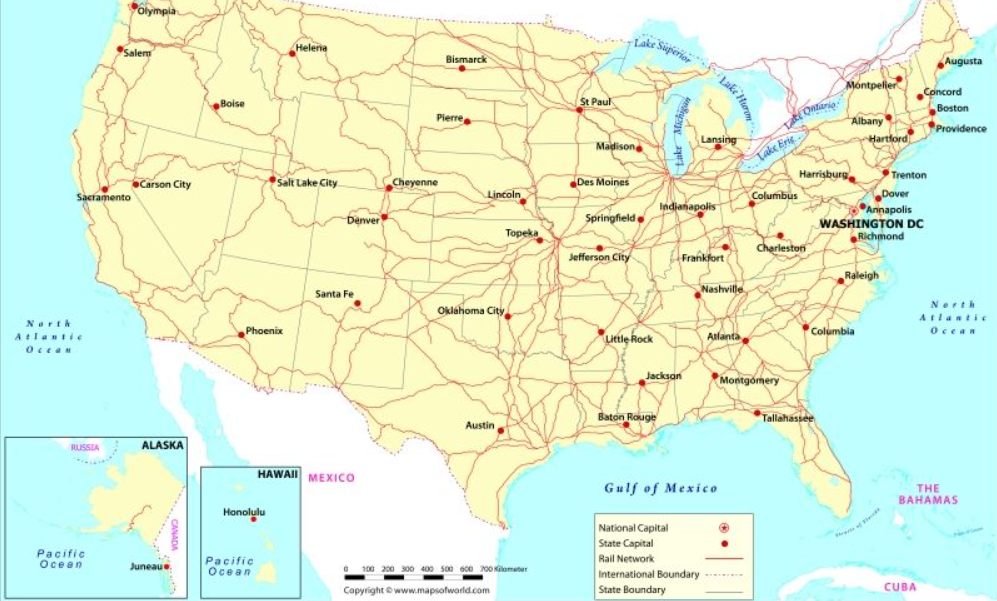}
    \caption{Strategic Locomotive Assignment: US railroad map \cite{mapsworld_usarail}.}
    \label{fig:fullnetworkslap}
\end{figure}

\subsection{Strategic Locomotive Assignment}
The second set of experiments considers the strategic locomotive assignment problem (SLAP)~\cite{kim2025practice}. The SLAP determines a weekly, repeatable plan for assigning locomotives to a fixed train schedule. The problem specifies the required inputs, operational constraints, and cost components governing locomotive movements and work events such as pickups and set-outs. Its main outputs are the weekly locomotive flows across the network and the locations where work events occur. These strategic decisions, including fleet sizing and event placement, provide the foundation for subsequent tactical and operational planning. The large problem setting includes 1{,}000 nodes and 40,000 arcs, and the very large (designated as \emph{-Hard}) problem setting includes 8{,}000 nodes and 50,000 arcs (see Figure~\ref{fig:fullnetworkslap} for an overview of the underlying US-wide railway network). In the large setting, instances reach optimality in an average runtime of five hours. In the very large setting, instances remain more demanding, and cannot close the gap even after 24 hours, with an average gap of 0.03\%.

In both sets of SLAP instances, the experiment assumes that the variation across problem instances is in the lower and upper bounds on the number of locomotives per arc. Because real fleets evolve with maintenance cycles, failures, seasonality, and acquisitions or retirements, the available number of locomotives is inherently time-varying. To generate instances, the experiments vary these bounds around reference values using a uniform distribution, ensuring the resulting limits remain realistic relative to operational constraints. Because the underlying US-wide railway network and the core train schedules remain completely static across instances, specific rail corridors retain a persistent identity. This structural consistency means that specific physical paths exhibit highly predictable, historically grounded utilization patterns. In this model, the general integer variables indicate the number of locomotives on each active arc and the number of light travel trains on each light travel arc. Given the vastness of the network, only a small subset of arcs is actually utilized, meaning the vast majority of these general integer variables are exactly zero. Identifying these zero-valued variables reveals the underlying sparsity of the physical routing, effectively pruning massive, unproductive regions of the decision space.

\begin{figure} [t]
    \centering
    \includegraphics[width=\linewidth]{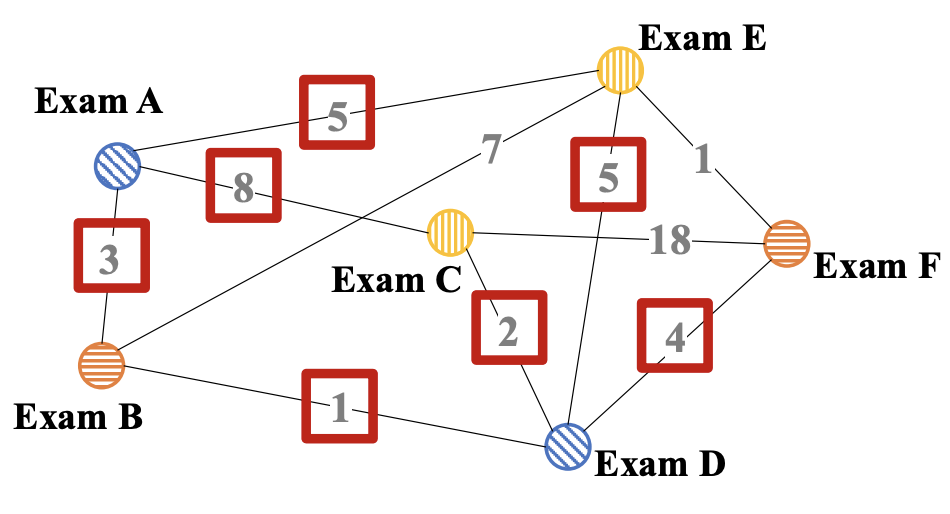}
    \caption{Course Scheduling: example from \cite{ye2025cornell}}
    \label{fig:fullnetworkslap}
\end{figure}

\subsection{Course Scheduling}
The third set of experiments considers a course scheduling problem (COURSE) based on the block sequencing formulation introduced in~\cite{ye2025cornell}. The COURSE determines an optimal assignment of university courses to discrete time blocks. The problem specifies the required inputs, operational constraints such as block capacities, and penalty components governing scheduling conflicts between co-enrolled courses. Its main outputs are the exact placements of courses into time slots. The large problem setting includes 200 courses and 12 time slots, and the very large (designated as \emph{-Hard}) problem setting includes 400 courses and 14 time slots. As the number of courses increases, the density of conflict constraints grows significantly, making the resulting MIPs substantially more challenging to solve.

We assumes that what varies across problem instances is the co-enrollment structure and associated penalty coefficients. Because real student enrollments fluctuate from term to term, the conflict density between courses is inherently time-varying. To generate instances, the experiments vary these parameters based on an anonymized version of Cornell University's Spring 2024 exam-scheduling input data using different random seeds, ensuring the resulting conflicts remain realistic relative to university operations. Despite these parameter variations, the physical dimensions of the problem (e.g. available time slots) remain perfectly fixed. This consistency ensures that individual courses maintain a distinct identity, exhibiting persistent historical scheduling characteristics regardless of term-to-term enrollment shifts. In this model, the discrete decisions are represented by binary variables indicating the assignment of each course to a specific time slot. Since each course must be placed into exactly one time slot out of all available options, all remaining assignment variables for that course are strictly zero. This structural sparsity highlights how isolating zero-valued binary variables can massively reduce the combinatorial search space.

\section{Experiments}

This section introduces the setup for empirical evaluation and presents the results. The code and instances will be provided open source upon acceptance.
%the instance files will be shared on the Distributional MIPLIB at \url{https://sites.google.com/usc.edu/distributional-miplib/}.

\begin{table}[t]
\centering
\caption{Overview of problem instances. 
\textbf{Columns:} \#Const.: number of constraints;  
\#Bin.: number of binary variables; \#Int.: number of general integer variables. }
\begin{tabular}{lrrr}
\toprule
Problem &  \#Const. & \#Bin. & \#Int. \\
\midrule
MMCNP       & 643      & 2{,}540  & 357  \\
MMCNP-hard  & 4{,}361  & 23{,}984 & 1{,}135 \\
\midrule
SLAP       & 80{,}982  & 8      & 77{,}946  \\
SLAP-hard  & 82{,}996  & 20     & 77{,}032  \\
\midrule
COURSE      & 30{,}173 & 43{,}488 & 173\\
COURSE-hard & 53{,}205 & 79{,}968 & 229 \\
\bottomrule
\end{tabular}
\label{tab:instances}
\end{table}
\begin{table*}[t]
\centering
\caption{Primal Gap (PG) and Primal Integral (PI) averaged over 100 test instances for each benchmark. We compare the performance of Gurobi, \textsc{PaS}, \textsc{PaS}+, and \textsc{ID-PaS+}. Results include the mean, standard deviation, and the number of instances each approach wins. The best-performing entries are highlighted in bold for clarity. We omit \textsc{PaS} for SLAP and SLAP-Hard benchmark because \textsc{PaS} is not applicable to instances with few binary variables.\\}
\begin{tabular}{c|c|ccc|ccc}
\toprule
                       &                  & \multicolumn{3}{c|}{PG (\%) $\downarrow$}                   & \multicolumn{3}{c}{PI  $\downarrow$}                      \\ \midrule
Benchmark             & Approach      & Mean            & Std Dev         & Wins        & Mean          & Std Dev       & Wins         \\ \midrule
\multirow{4}{*}{MMCNP}  & Gurobi           & 0.18          & 0.25          & 9           & 3.50         & 3.28          & 9            \\
                       & \textsc{PaS}  & 0.16 (11.1\%) & 0.25 & 15 & 3.49 (0.29\%) & 2.71 & 7 \\
                       & \textsc{PaS}+      & 0.14 (22.2\%)          & 0.25 & 23          & 2.42 (30.9\%)          & 2.53         & 19           \\
                       & \textsc{ID-PaS+} & \textbf{0.09 (50.0\%)} & 0.18        & \textbf{53} & \textbf{2.13 (39.1\%)} & 2.06 & \textbf{65}  \\ \midrule
\multirow{4}{*}{MMCNP-Hard} & Gurobi            & 1.07          & 0.49          & 2          & 35.37         & 5.83          & 0            \\
                    & \textsc{PaS} & 0.58 (45.8\%) & 0.40 & 10 & 18.28 (48.3\%) & 4.82 & 9 \\
                       & \textsc{PaS}+      & 0.40 (62.6\%)     & 0.39          & 21 & 16.10 (54.5\%) & 4.00          & 15  \\
                       & \textsc{ID-PaS+} & \textbf{0.15 (86.0\%)} & 0.25 & \textbf{67}          & \textbf{11.53 (67.4\%)}          & 2.67 & \textbf{76}           \\ \midrule
\multirow{3}{*}{SLAP} & Gurobi           & 0.049           & 0.054 & 15          & 1.36          & 0.59         & 13            \\
                       & \textsc{PaS}+      & 0.025 (49.0\%)           & 0.036         & 31 & 1.30 (4.4\%) & 0.50 & 20          \\
                       & \textsc{ID-PaS+} & \textbf{0.024 (51.0\%)}  & 0.032          & \textbf{54}          & \textbf{1.16 (14.7\%)} & 0.43         & \textbf{67}  \\ \midrule               
\multirow{3}{*}{SLAP-Hard} & Gurobi            & 0.18          & 0.21         & 18          & 7.85         & 2.12          & 13            \\
                       & \textsc{PaS}+      & 0.16 (11.1\%)          & 0.19          & 21          & 6.36 (19.0\%)          & 1.90          & 14           \\
                       & \textsc{ID-PaS+} & \textbf{0.13 (27.8\%)} & 0.15 & \textbf{61} & \textbf{5.36 (31.7\%)} & 1.55 & \textbf{73}  \\ \midrule
\multirow{4}{*}{COURSE} & Gurobi            & 0.023           & 0.010 & 2          & 1.97          & 1.78        & 21          \\
                        & \textsc{PaS} &  0.013 (43.5\%) & 0.008 & 18 & 1.90 (3.6\%) & 1.59 & 25\\
                       & \textsc{PaS}+      & 0.024 (-4.4\%)           & 0.012         & 1 & 2.08 (-5.6\%) & 1.82 & 13         \\
                       & \textsc{ID-PaS+} & \textbf{0.004 (82.6\%)}  & 0.003          & \textbf{79}          & \textbf{1.75 (11.2\%)} & 1.71         & \textbf{41}  \\ \midrule               
\multirow{4}{*}{COURSE-Hard} & Gurobi            & 0.26         & 0.44         & 3          & 9.87         & 6.35          & 11            \\
                    & \textsc{PaS} & 0.13 (50.0\%) & 0.31 & 6 & 9.14 (7.4\%) & 5.03 & 9\\
                       & \textsc{PaS}+      & 0.10 (61.5\%)          & 0.29          & 17          & 8.38 (15.1\%)          & 5.05         & 15           \\
                       & \textsc{ID-PaS+} & \textbf{0.07 (73.1\%)} & 0.23 & \textbf{74} & \textbf{7.61 (22.9\%)} & 4.56 & \textbf{65}  \\ \bottomrule
\end{tabular}
\label{table: pas}
\end{table*}

\subsection{Experiment Setup}

\paragraph{\textbf{Baselines:}}
We evaluate the performance of \textsc{ID-PaS+} against the following three baselines:
\begin{enumerate}
    \item \textbf{Gurobi (v12.0.3)}~\cite{gurobi}: A state-of-the-art commercial MIP solver, run with default parameter settings and all built-in primal heuristics enabled. This serves as a strong industrial reference, reflecting standard solver performance without learning-based guidance.
    \item \textbf{\textsc{PaS}}~\cite{han2023gnn}: The original Predict-and-Search framework, which attempts to predict both zero and one assignments exclusively for binary variables.
    \item \textbf{\textsc{PaS}+}: An extended version of \textsc{PaS} adapted to support general MIP domains by predicting solely zero-valued variables. Because this baseline lacks identity-aware positional embeddings, it serves as a crucial ablation to isolate the performance gains of our zero-prediction mechanism from the benefits of identity awareness.
\end{enumerate}

For all ML-based approaches (\textsc{PaS}, \textsc{PaS}+, and \textsc{ID-PaS+}), a separate model is trained for each benchmark domain and difficulty setting. Each model is trained on 480 instances, validated on 120 instances, and evaluated on a disjoint set of 100 test instances. Summary statistics for the reference instances are reported in Table~\ref{tab:instances}.

\begin{figure*}[t!]

 \centering
 \begin{subfigure}{0.32\textwidth}
     \includegraphics[width=\textwidth]{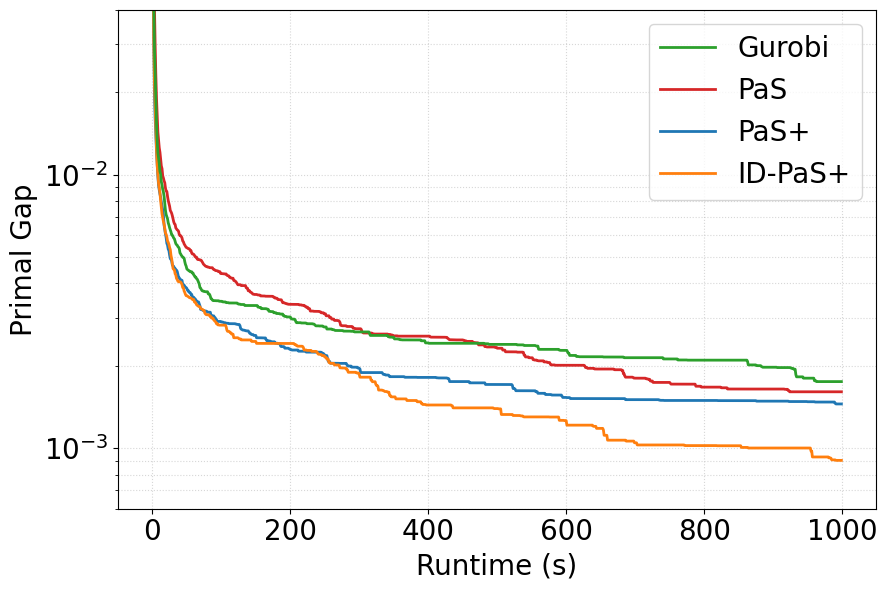}
     \caption{MMCNP}
 \end{subfigure}
 \hfill
  \begin{subfigure}{0.32\textwidth}
     \includegraphics[width=\textwidth]{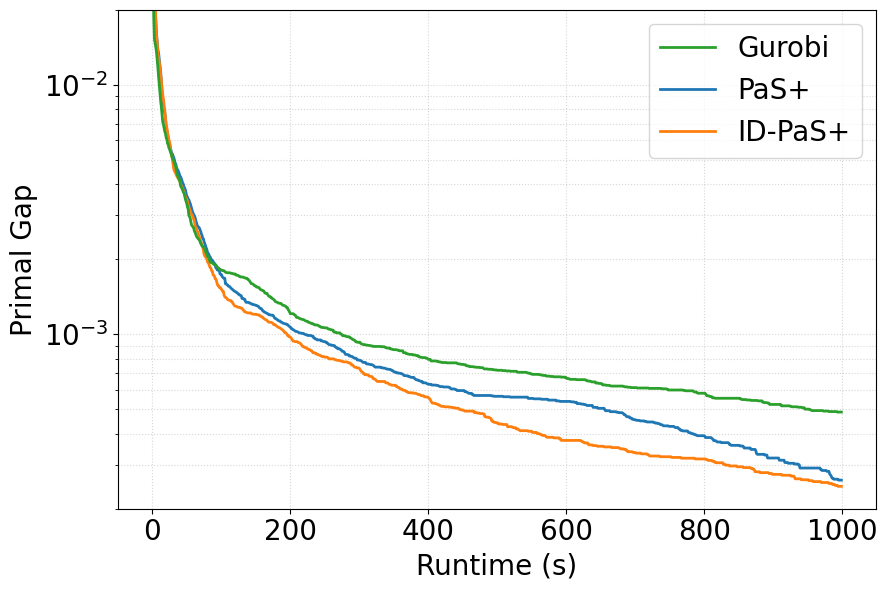}
     \caption{SLAP}
 \end{subfigure}
 \hfill
  \begin{subfigure}{0.32\textwidth}
     \includegraphics[width=\textwidth]{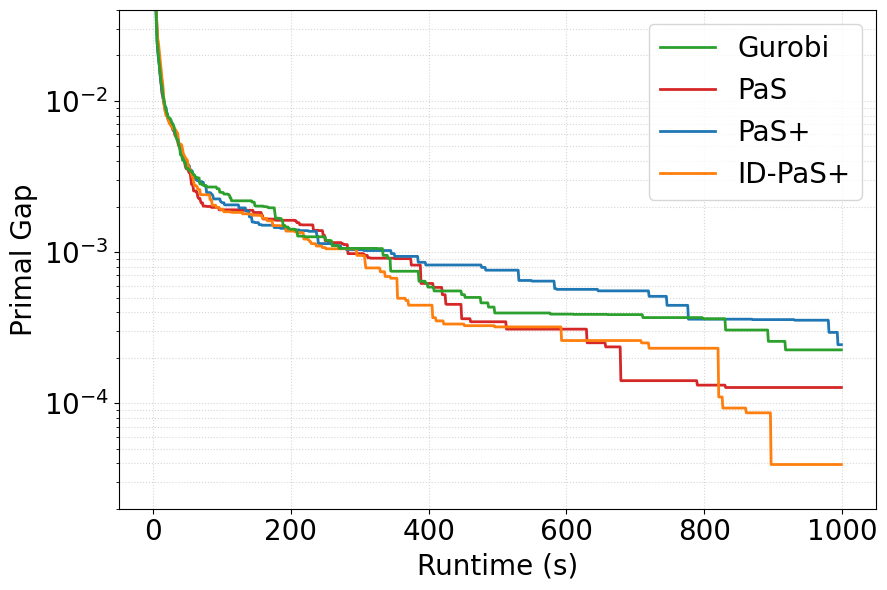}
     \caption{COURSE}
 \end{subfigure}
 \begin{subfigure}{0.32\textwidth}
     \includegraphics[width=\textwidth]{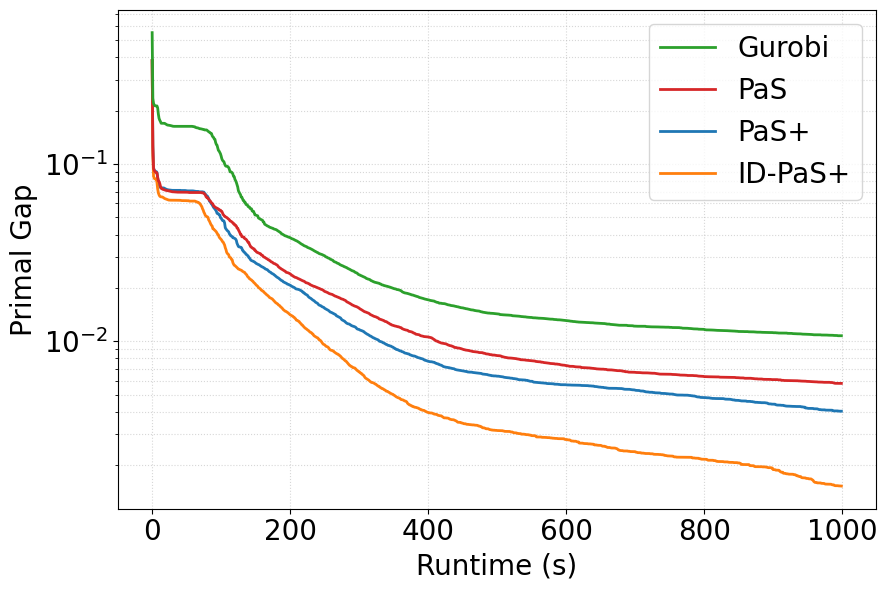}
     \caption{MMCNP-Hard}
 \end{subfigure}
 \hfill
 \begin{subfigure}{0.32\textwidth}
     \includegraphics[width=\textwidth]{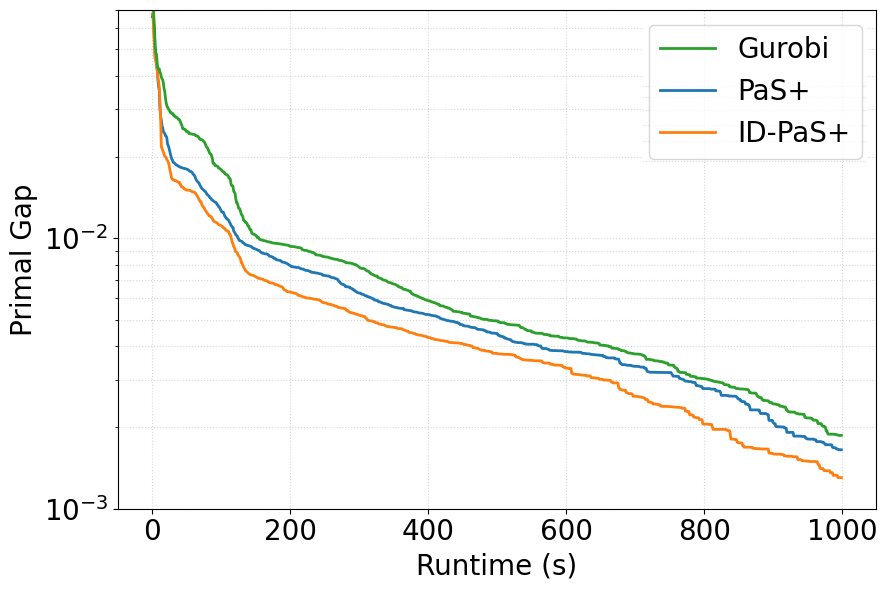}
     \caption{SLAP-Hard}
 \end{subfigure}
 \hfill
  \begin{subfigure}{0.32\textwidth}
     \includegraphics[width=\textwidth]{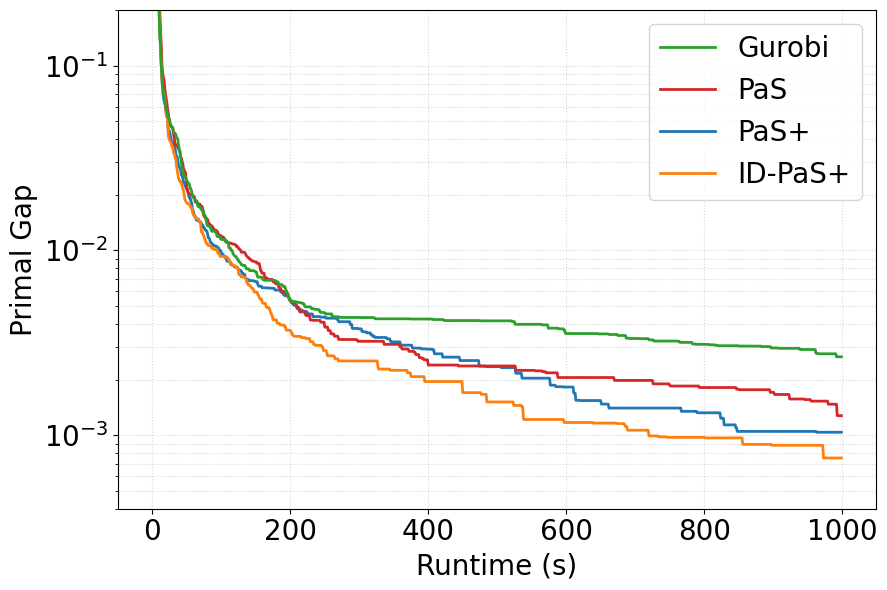}
     \caption{COURSE-Hard}
 \end{subfigure}
 \caption{The Primal Gap (the lower, the better) over time, averaged over 100 test instances on each benchmark. We compare Gurobi (green line), \textsc{PaS} (red line), \textsc{PaS}+ (blue line), and \textsc{ID-PaS+} (orange line). We omit \textsc{PaS} for SLAP and SLAP-Hard benchmark because \textsc{PaS} is not applicable to instances with few binary variables.}
 \label{fig: pas}
\end{figure*}

\paragraph{\textbf{Evaluation Metrics:}}
We evaluate solver performance on the test instances using two primary metrics:
\begin{enumerate}
    \item \textbf{Primal Gap (PG)}~\cite{berthold2006primal}: Defined as the normalized difference between the primal bound $v$ and the best-known objective value $v^*$: 
    $
    \text{PG} = \frac{|v - v^*|}{\max(|v^*|, \epsilon)}
    $
    where $\epsilon = 10^{-8}$ is used to avoid division by zero. This metric is computed when a feasible solution $v$ exists and $v \cdot v^* \geq 0$. 
    \item \textbf{Primal Integral (PI)}~\cite{achterberg2012rounding}: The integral of the primal gap over the runtime interval $[0, t]$. The PI provides a comprehensive measure of both the quality of the solutions found and the speed at which they are discovered.
\end{enumerate}

\paragraph{\textbf{Hyperparameters:}}
All experiments are conducted on 2.4 GHz Intel Xeon-2640v3 CPUs with 64 GB of memory. Gurobi v12.0.3 is used as the underlying MIP solver for data collection and the evaluation of all approaches. All solver runtimes during evaluation are strictly capped at 1{,}000 seconds. Neural network training is accelerated using an NVIDIA V100 GPU with 112 GB of memory. Models are trained using the Adam optimizer~\cite{kingma2014adam} with a learning rate of $1 \times 10^{-5}$ and a batch size of 16 for a maximum of 48 hours. The model checkpoint achieving the lowest validation loss is selected for testing. 

After training, we perform hyperparameter tuning for the search phase on a subset of 30 validation instances via a grid search. For the \textsc{PaS} baseline, the neighborhood parameters $(k_0, k_1, \Delta)$ are tuned , with $\Delta \in \{1\%, 3\%, 5\%\}$ and the target binary variable prediction parameters $k_0, k_1 \in \{0\%, 10\%, \dots, 50\%\}$. For both \textsc{PaS}+ and \textsc{ID-PaS+}, the parameters $(k_0^b, k_0^i, \Delta)$ are tuned on the same 30 validation instances. Specifically, we tune with $\Delta \in \{1\%, 3\%, 5\%\}$ and the target prediction percentages $k_0^b$ and $k_0^i$ across $\{0\%, 10\%, 30\%, \dots, 90\%\}$ of their respective variable counts. This adaptive tuning strategy ensures that the framework can robustly handle general MIPs, as well as problems containing exclusively binary or integer variables.

\subsection{Results}

Table~\ref{table: pas} and Figure~\ref{fig: pas} demonstrate that \textsc{ID-PaS+} achieves substantial, state-of-the-art improvements over default Gurobi, the original \textsc{PaS} framework, and the \textsc{PaS}+ ablation baseline across 100 previously unseen test instances for each benchmark. A Wilcoxon signed-rank test ($p < 0.01$)~\cite{woolson2007wilcoxon} confirms that all reported performance differences between \textsc{ID-PaS+} and the baselines are statistically significant, validating the robustness of the observed improvements across varying difficulty levels.

\paragraph{\textbf{The Impact of Zero-Prediction (\textsc{PaS}+ vs. \textsc{PaS}):}}
Table~\ref{table: pas} demonstrates the need to extend the Predict-and-Search (\textsc{PaS}) framework beyond binary classifications. While the original \textsc{PaS} improves upon Gurobi on MMCNP and COURSE, it fails on SLAP benchmarks due to its inability to handle general integers. By restricting the learning target strictly to zero-valued variables, \textsc{PaS}+ bridges this gap. This modification yields average gains of roughly 30\% on Primal Gap (PG) and 12\% on Primal Integral (PI) over Gurobi across SLAP and SLAP-Hard. Furthermore, \textsc{PaS}+ outperforms the original \textsc{PaS} on the MMCNP, MMCNP-Hard, and COURSE-Hard benchmarks by an additional 13\% on PG and 15\% on PI on average. Ultimately, the improvements on MMCNP demonstrate that predicting general integer variables is crucial for performance. Furthermore, while there is a slight performance dip on the COURSE, the overall results indicate that focusing the solver's trust region exclusively around highly confident zero-predictions does not broadly harm performance, and in more complex scenarios, significantly improves it.

\paragraph{\textbf{The Necessity of Identity-Awareness (\textsc{ID-PaS+} vs. PaS+):}}
While \textsc{PaS}+ establishes a strong baseline, \textsc{ID-PaS+} consistently and significantly outperforms it across every single benchmark and difficulty setting. This performance gap isolates the immense value of identity-aware positional embeddings. By allowing the network to recognize specific physical infrastructure (e.g., persistent rail corridors or fixed highway arcs) rather than treating nodes as anonymous entities, \textsc{ID-PaS+} provides far more accurate zero-predictions. Overall, \textsc{ID-PaS+} achieves the lowest Primal Gap (PG)  and Primal Integral (PI) across the board, reducing PG by 27.8\% to 86.0\% and PI by 11.2\% to 67.4\% relative to Gurobi, while winning the absolute majority of individual test instances.

\begin{figure}[h]
\centering
\begin{subfigure}{0.45\textwidth}
    \includegraphics[width=0.9\linewidth]{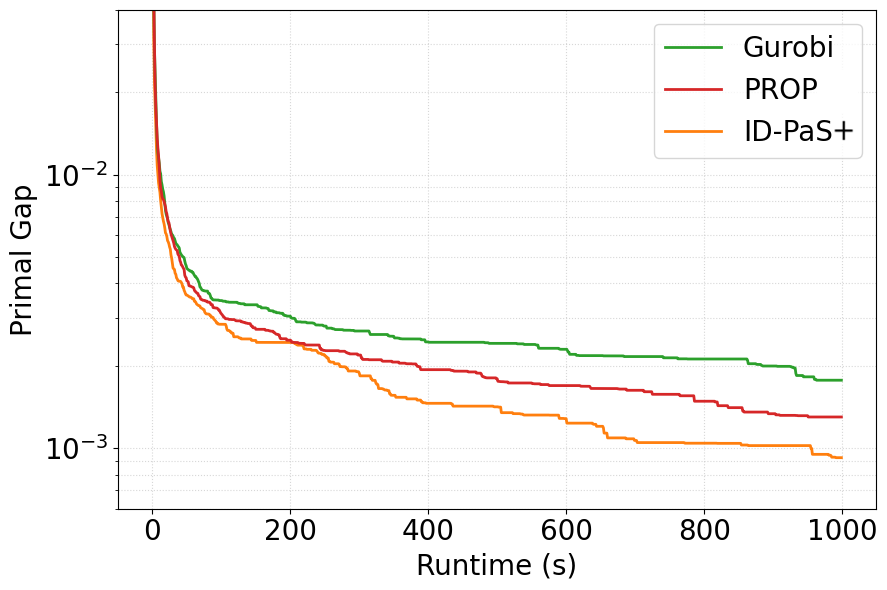}
\end{subfigure}
\begin{subfigure}{0.45\textwidth}
    \includegraphics[width=0.9\linewidth]{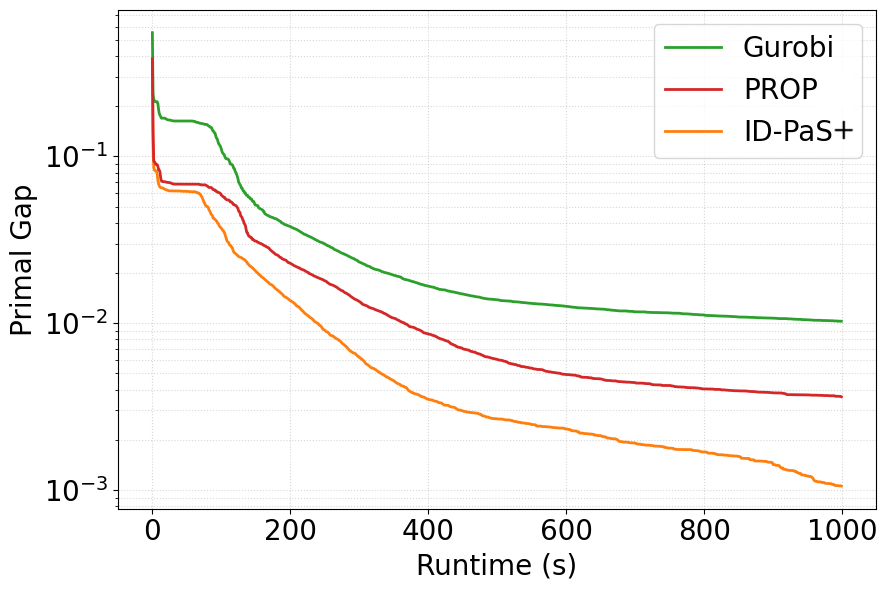}
\end{subfigure}
\caption{Primal gap (lower is better) over time, averaged over 100 test instances on MMCNP (top) and MMCNP-Hard (bottom). We compare Gurobi (green), PROP (red), and \textsc{ID-PaS+} (orange).}
\label{fig:propel}
\end{figure}

\paragraph{\textbf{Runtime Trajectory and Anytime Performance:}}
Figure~\ref{fig: pas} illustrates the anytime performance improvements of our framework over the 1000-second evaluation horizon. On the easier benchmarks, while the advantages of \textsc{ID-PaS+} are less pronounced during the initial 100 to 200 seconds, \textsc{ID-PaS+} continues to aggressively reduce the primal gap at later runtimes, even after \textsc{PaS} and \textsc{PaS}+ have plateaued. On the harder benchmarks, the performance gap is even more pronounced: \textsc{ID-PaS+} identifies significantly higher-quality solutions almost immediately upon initialization and consistently widens its lead over \textsc{PaS}, \textsc{PaS}+, and Gurobi throughout the entire time limit. These anytime performance plots clearly demonstrate that identity-aware zero-predictions provide highly reliable guidance and large improvement for large-scale, real-world MIPs.

\paragraph{Hyperparameters for search:} 
Across all benchmarks, the best-performing hyperparameter configurations fix approximately 70\% or 90\% of the variables, utilizing a trust region $\Delta = 1\%$ of $|X_0|$. We find that for benchmarks, where a single variable type heavily dominates (Binary for COURSE and Integer for SLAP, see Table~\ref{tab:instances}), focusing exclusively on predicting that specific type yields the best performance during hyperparameter tuning. In the MMCNP domain, we surprisingly find that the best-performing configuration only predicts integer variables, although they are much fewer than the binary variables, indicating their outsized importance in this specific problem structure. Ultimately, the adaptive hyperparameter search across both variable types equips the framework to handle diverse problem scenarios robustly.

\paragraph{\textbf{Comparison Against Static Heuristics (Zero-Frequency):}}
To further validate the need for graph-learning methods, we compare \textsc{ID-PaS+} against a baseline that fixes the same number of $|X_0|$ variables based solely on their empirical probability of being exactly zero in the training data. As shown in Table~\ref{tab:ablation_zerofreq}, \textsc{ID-PaS+} dominates this baseline across all datasets. By leveraging contextual graph features rather than historical averages, \textsc{ID-PaS+} wins the vast majority of the 100 test instances.

\begin{table}[t]
\centering
\caption{Comparison of \textsc{ID-PaS+} against the Zero-Frequency Baseline. Values represent the number of instances (out of 100) where \textsc{ID-PaS+} achieved a strictly better Primal Gap (PG) or Primal Integral (PI).}
\label{tab:ablation_zerofreq}
\begin{tabular}{@{}lcc@{}}
\toprule
\textbf{Benchmark} & \textbf{\textsc{ID-PaS+} Wins (PG)} & \textbf{\textsc{ID-PaS+} Wins (PI)} \\ \midrule
MMCNP              & 83                                 & 71                                 \\
SLAP               & 76                                 & 85                                 \\
COURSE             & 82                                 & 76                                 \\ \bottomrule
\end{tabular}

% \begin{tabular}{@{}lcc@{}}
% \toprule
% \textbf{Benchmark} & \textbf{\textsc{PaS+} Wins (PG)} & \textbf{\textsc{PaS+} Wins (PI)} \\ \midrule
% MMCNP              & 68                                 & 64                                 \\
% SLAP               & 70                                 & 81                                 \\
% COURSE             & 72                                 & 63                                 \\ \bottomrule
% \end{tabular}
\end{table}

\begin{table}[h]
\centering
\caption{Instance-wise win statistics (out of 100) comparing Gurobi, PROP, and \textsc{ID-PaS+} on the MMCNP benchmarks. Remaining instances are won by Gurobi.}
\label{tab:ablation_prop}
\begin{tabular}{@{}llcc@{}}
\toprule
\textbf{Benchmark} & \textbf{Approach} & \textbf{Wins (PG)} & \textbf{Wins (PI)} \\ \midrule
\multirow{2}{*}{MMCNP}      & PROP             & 16                 & 18                 \\
                            & \textsc{ID-PaS+} & \textbf{59}        & \textbf{62}        \\ \midrule
\multirow{2}{*}{MMCNP-Hard} & PROP             & 21                 & 7                  \\
                            & \textsc{ID-PaS+} & \textbf{63}        & \textbf{78}        \\ \bottomrule
\end{tabular}
\end{table}

\paragraph{\textbf{Comparison Against Independent Predictions (PROP):}}
Finally, we implement the PROP framework from PROPEL~\cite{akhlaghi2025propel}, a strong learning baseline that is inherently identity-aware. PROP trains a dedicated neural network for each individual variable to predict its assignment independently. Because training tens of thousands of individual models is computationally demanding for the massive variable counts in SLAP and COURSE, we evaluate PROP exclusively on the MMCNP and MMCNP-Hard benchmarks.

Figure~\ref{fig:propel} and Table~\ref{tab:ablation_prop} demonstrate that while PROP improves upon default Gurobi by using a variable-specific approach, \textsc{ID-PaS+} consistently secures the majority of wins on both Primal Gap and Primal Integral. Although PROP successfully captures distinct variable identities by dedicating a model to each variable, the performance gap indicates that identity awareness alone is insufficient. \textsc{ID-PaS+}, which leverages a shared graph-based representation augmented with identity-aware embeddings, is necessary to achieve the strongest overall performance.

\section{Conclusion}

This work introduces \textsc{ID-PaS+} to extend the Predict-and-Search framework to general Mixed-Integer Programs (MIPs). \textsc{ID-PaS+} leverages a zero-prediction mechanism to exploit structural sparsity and incorporates identity-aware embeddings to capture persistent physical topologies across varying instances. Extensive experiments on large-scale, real-world benchmarks demonstrate that \textsc{ID-PaS+} consistently outperforms default Gurobi, the original binary-only \textsc{PaS}, and independent per-variable baselines like PROP. By effectively modeling global structural dependencies, \textsc{ID-PaS+} achieves up to an 86.0\% reduction in Primal Gap and a 67.4\% reduction in Primal Integral relative to the state-of-the-art solver, offering a highly reliable and general approach for accelerating complex optimization tasks.

\section{Acknowledgment}
The National Science Foundation (NSF) partially supported the research under grant \#2112533:"NSF Artificial Intelligence Research Institute for Advances in Optimization (AI4OPT)" and grant \#2346058: "NRT-AI: Integrating Artificial Intelligence and Operations Research Technologies".

\bibliography{ref}

% Check whether the conference requires a reproducibility checklist to be included in the paper.
% If so, you can uncomment the following line and ajust the path to include it.
% \input{../../ReproducibilityChecklist/LaTeX/ReproducibilityChecklist.tex}

\end{document}